\documentclass[a4paper,12pt]{article}

\usepackage[margin=25mm]{geometry}
 \usepackage{graphicx}
 \usepackage{xcolor}
 \usepackage{verbatim}
 \usepackage{multicol,multirow}
 \usepackage[utf8]{inputenc}
 \usepackage{amsmath,amsthm} \allowdisplaybreaks
 \usepackage{amssymb}
 \usepackage{array}
 \usepackage{booktabs}
\usepackage{caption}
\usepackage{mathtools}
\usepackage{csvsimple}
\usepackage{subcaption}
\usepackage{hyperref}
\usepackage[numbers]{natbib}
\usepackage{algorithm}
\usepackage{algorithmic}

\usepackage{booktabs}
\usepackage{multirow}
\usepackage{amssymb}
\usepackage{amsmath}
\usepackage{microtype}
\usepackage{subcaption}
\usepackage{graphicx}
\usepackage[inline]{enumitem}

\usepackage[dvipsnames]{xcolor}
\usepackage[draft,nomargin,inline]{fixme}
\fxsetup{theme=color,mode=multiuser}
\FXRegisterAuthor{mh}{amh}{\color{RoyalBlue}MH}
\FXRegisterAuthor{mll}{amll}{\color{Green}Mimi}
\FXRegisterAuthor{pm}{apml}{\color{Red}Patrick}

\newcommand{\eqfontsize}{\small \selectfont}

\date{April 2, 2025}

\title{Solving Parallel Machine Scheduling \\ With Precedences and Cumulative Resource Constraints With Calendars}

\author{Christoph Einspieler$^{2}$,
Matthias Horn$^{2}$, 
Marie-Louise Lackner$^{1}$, \\
Patrick Malik$^{2}$, 
Nysret Musliu$^{1}$, 
Felix Winter$^{1}$\\
\\
\small \texttt{christoph.einspieler@mcp-alfa.com}\\
\small \texttt{matthias.horn@mcp-alfa.com}\\
\small \texttt{marie-louise.lackner@tuwien.ac.at}\\
\small \texttt{patrick.malik@mcp-alfa.com}\\
\small \texttt{nysret.musliu@tuwien.ac.at}\\
\small \texttt{felix.winter@tuwien.ac.at}\\
\\
 $^{1}$\small\emph{Databases and Artificial Intelligence Group} (DBAI), TU Wien, Vienna,  Austria \\
 $^{2}$\small \emph{MCP GmbH}, Vienna, Austria \\
}

\begin{document}

\maketitle

\begin{abstract}
The task of finding efficient production schedules for parallel machines is a challenge that arises in most industrial manufacturing domains.
There is a large potential to minimize production costs through automated scheduling techniques, due to the large-scale requirements of modern factories.
In the past, solution approaches have been studied for many machine scheduling variations, where even basic variants have been shown to be NP-hard.
However, in today's real-life production environments,  additional complex precedence constraints and resource restrictions with calendars arise that must be fulfilled.
These additional constraints cannot be tackled efficiently by existing solution techniques.
Thus, there is a strong need to develop and analyze automated methods that can solve such real-life parallel machine scheduling scenarios.

In this work, we introduce a novel variant of parallel machine scheduling with job precedences and calendar-based cumulative resource constraints that arises in real-life industrial use cases.
A constraint modeling approach is proposed as an exact solution method for small scheduling scenarios together with state-of-the-art constraint-solving technology.
Further, we propose a  construction heuristic as well as a tailored metaheuristic using local search to efficiently tackle large-scale problem instances.
This metaheuristic approach has been deployed and is currently being used in an industrial setting.

To evaluate the proposed techniques, we provide a set of benchmark instances consisting of real-life and randomly generated scheduling scenarios.
The experimental results show that the constraint modeling approach can provide optimal results for the majority of small instances but cannot efficiently tackle large-scale problems. 
However, the proposed heuristic approaches successfully provide high-quality results within short run-time limits for all large-scale real-life scheduling scenarios and almost all random instances.
\end{abstract}

\section{Introduction}

Modern industrial factories use sophisticated parallel machine environments to manufacture products on a large scale.
Thus, creating efficient production schedules that fulfill all requirements in the usually highly automated process is often a difficult task, as many complex constraints have to be considered while a multi-objective cost function needs to be minimized.
To address this challenge, automated scheduling techniques have been successfully applied to create optimized schedules and have been shown to significantly enhance the quality of practical production planning. 
Although solution approaches have been previously studied for a variety of machine scheduling variants, there are many industrial applications that require the consideration of unique constraints and objectives that cannot be tackled efficiently by existing methods.
Thus, there is still a strong need to develop new automated solution approaches to ensure adequate and cost-efficient schedules under such new conditions.

In this work, we study a new variant of the parallel machine scheduling problem (PMSP) that includes several unique restrictions, such as calendar-based cumulative resource constraints, complex precedence requirements between jobs, and machine calendars.
PMSPs have been widely studied in the literature. Since a basic variant involving just two machines was proven to be NP-complete~\cite{gary1979computers}, numerous extensions with additional parameters have been explored.
A recent survey on the topic~\cite{durasevic_heuristic_2023} provides a detailed overview of previous work and describes many different constraints and objectives that arise with PMSPs.

Investigating relevant practical problem variants remains an important and active field of research, as more complex scheduling tasks keep arising in the industry due to the ongoing trend towards full automation.
For example, the complexity of a PMSP variant in which pairs of incompatible jobs cannot be scheduled on the same machine was recently investigated~\cite{pikies_scheduling_2021}.
Another recently studied variant of the problem considers a multi-objective function that aims to minimize both the makespan and the energy consumption and has been successfully tackled using a construction heuristic together with a local search procedure~\cite{anghinolfi_bi-objective_2021}.
The variant of PMSP investigated in this paper has a multi-criteria objective as well. It aims to minimize a weighted linear combination of multiple objectives: makespan, total tardiness, and total job setup time. While related variants of PMSP have been discussed in the literature---some of which share certain characteristics with the version considered here---none of the existing problem formulations or solution approaches can be directly applied to our variant. 
More specifically, Dang et al.~\cite{dang_matheuristic_2021} propose a hybrid approach combining Mixed-Integer Programming (MIP) and Genetic Algorithms to solve a PMSP variant that minimizes tardiness and setup times. However, their method is limited to a restricted model of tool set resources, making it inapplicable to PMSPs with general resource constraints. A related variant that incorporates multi-resource requirements and job precedence constraints is addressed by Yunusoglu et al.~\cite{yunusoglu_constraint_2022} using a Constraint Programming (CP) approach. Nevertheless, this method does not support calendar-based machine and resource constraints and defines the makespan as the optimization objective. Although calendar-based cumulative resource constraints have been explored in the context of Resource-Constrained Project Scheduling (RCPS), for example by Kreter et al.~\cite{DBLP:journals/constraints/KreterSS17}, the PMSP variant examined in this paper exhibits several distinctive features. These include specific machine resource requirements and resource consumption during machine downtimes, which are not typically addressed in RCPS formulations. Moser et al.~\cite{moser_exact_2022} propose a metaheuristic local search-based approach using simulated annealing and several MIP models to tackle a PMSP with makespan and tardiness objectives. However, this formulation does not account for resource or precedence constraints.
Lastly, a PMSP with machine availability constraints has been solved by Santoro et al.~\cite{santoro_unrelated_2023} using an exact solution method based on MIP, but the approach does not model resource or precedence constraints.

In this work, we introduce a new PMSP variant originating from a real-life scheduling application in the packaging industry. The studied problem, which we refer to as the\textit{Parallel Machine Scheduling Problem with Precedences and Cumulative Resource Constraints With Calendars} (PMSP-PCRCC), presents significant challenges as it subsumes unique calendar-based resource restrictions together with and a diverse set of scheduling objectives and structural constraints.  
To the best of our knowledge, prior work has addressed only subsets of these requirements.
Furthermore, the PMSP-PCRCC uses a more general notion of job precedence constraints than usual in the related literature, where jobs can define multiple predecessor requirements and specify a minimum distance to their predecessors.
Thus, new solution methods are required to efficiently tackle all traditional requirements together with novel calendar-based resource- and job-precedence constraints.

The main contributions of this paper are summarized as follows:
\begin{itemize}
\item We formally introduce a novel PMSP variant originating from the packaging industry.
\item We provide 25 real-life instances and a large set of random instances as well as a random instance generator.
\item We propose a solver-independent constraint modeling approach that can be used with state-of-the-art constraint-solving technology as an exact solution method. The approach can successfully provide optimal solutions for the large majority of the small random instances.
\item We develop a metaheuristic approach using local search to tackle large-scale real-life scheduling scenarios. Thereby, we present an innovative solution representation and search neighborhood to efficiently handle the problem's unique resource and precedence constraints.
This approach is successfully being used in an industrial setting within the packaging industry.
\item A construction heuristic is proposed that can quickly produce solutions for large instances and thereby provide efficient initial solutions to the metaheuristic.
\item We evaluated all proposed methods empirically and show that the methods are able to reach optimal results for many instances and provide high-quality solutions for all large-scale realistic instances.
\end{itemize}

In the following, we first introduce the problem specification in Section~\ref{sec:problem-def}. Afterward, the constraint modeling approach is described in Section~\ref{sec:cp_model}. Details about the construction heuristic and metaheuristic methods are given in Sections~\ref{sec:construction_heuristic} and~\ref{sec:sim-ann}.
The experimental results are presented in Section~\ref{sec:experiments} before final concluding remarks are made.

\section{Problem Definition}
\label{sec:problem-def}

The real-life environment from which our problem originates requires a complex parallel machine setup. Thus, many jobs need to be scheduled on a set of machines, where each job can only be assigned to a subset of eligible machines and defines machine-dependent processing times.

\begin{figure}[ht]
  \centering
  \includegraphics[width=\columnwidth]{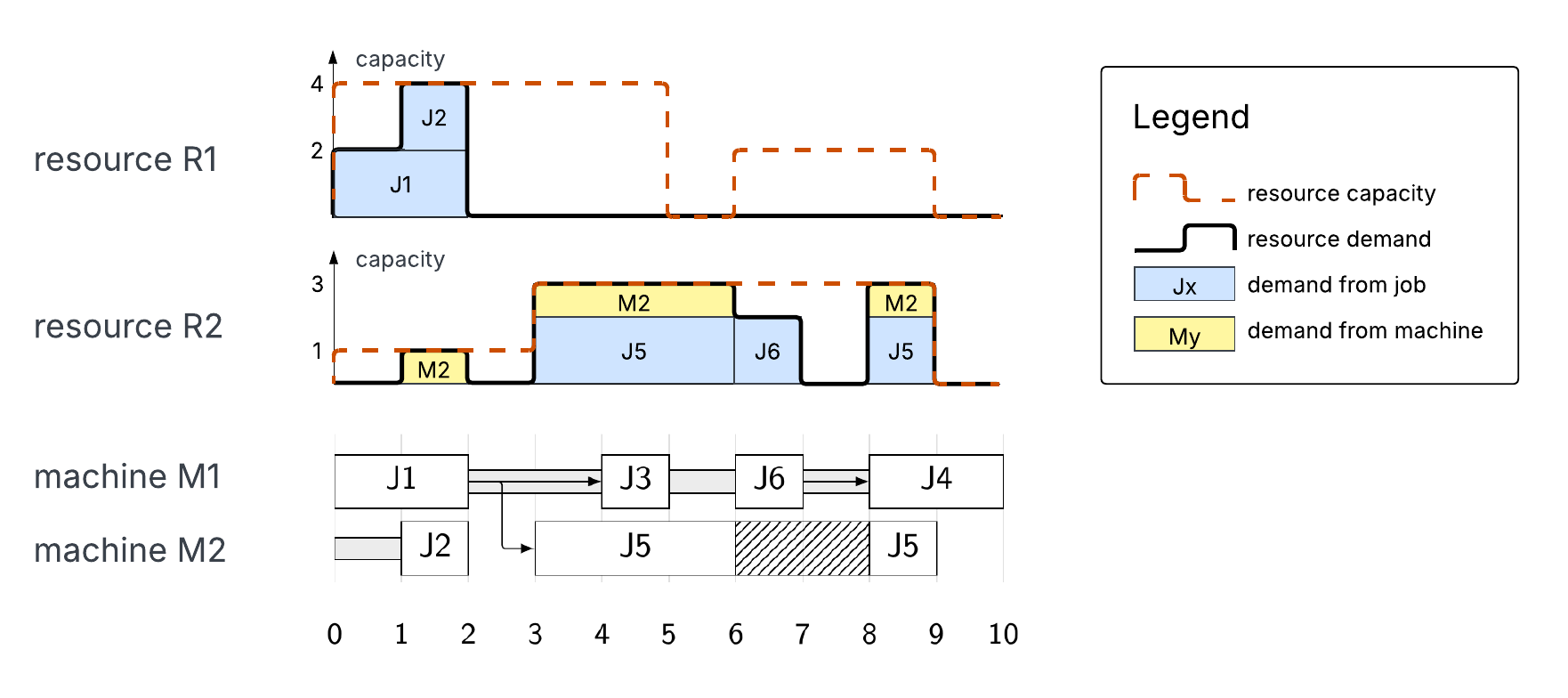}
  \caption{An example schedule with two parallel machines, two resources and six jobs.}\label{fig:example_schedule}
\end{figure}

Figure~\ref{fig:example_schedule} illustrates a simple example schedule using two machines (M1, M2), two resources (R1, R2), and six jobs (J1-J6) in the form of a resource consumption chart (upper part of the figure) and a Gantt chart (lower part of the figure).
In the Gantt chart, jobs are drawn as white bars, where the horizontal length indicates the processing time. The light gray bars with smaller widths denote the predecessor-dependent setup time of the jobs. Note that there is a single case with a 0-length setup time (setup between J2 and J5).

The diagonally striped bar within the schedule of M2 indicates a machine downtime period. During a downtime, jobs cannot be processed. However, jobs running before a downtime period can simply be paused. In the example, this is illustrated as J5 is interrupted but continues immediately after the end of the downtime.

In any feasible schedule, job precedence constraints have to be respected. In the example, job precedences are illustrated by arrows between pairs of jobs. Thus, J3 and J5 must be scheduled after the completion of J1, and J4 needs to bes scheduled after the completion of J6. Precedences can optionally specify minimum time lags between jobs. In the example, J5 must respect a short time lag of one time unit after the completion of J1 (otherwise, J5 could start directly after J2 on M2).

In addition, two resources (R1, R2) are visualized in the upper part of Figure~\ref{fig:example_schedule}. Cumulative resource constraints impose restrictions on feasible schedules as follows: For each time slot in the scheduling horizon, each resource provides a certain integer-valued amount of capacity. Thus, resource calendars can be modeled by supplying different capacities in different time periods of the scheduling horizon.
Each job also specifies resource demands that must be fulfilled. Moreover, machine resource demands can be defined (in other words, a machine might require certain resources while processing any job).
Thus, for each time slot, all resources need to provide sufficient capacity to supply all jobs that are processed and all machines that are processing some job at the given time. 
Note that jobs paused during machine downtimes do not consume any resource capacities. Job preemption is however not allowed during machine runtimes, i.e., jobs may not be paused outside of machine downtimes to purposely interrupt resource consumption.

In the example, resource capacities are indicated by a bold dashes line:  R1 provides a capacity of 4 in the interval [0,5) and a capacity of 2 in the interval [6,9); R2 provides a capacity of 1 during the interval [0,3) and a capacity of 3 during the interval [3,9); during the remaining times, there is no resource capacity supply.
Note that every job and every machine can require different sets of resources and that the required demands can vary in magnitude. 
In the example, J1 and J2 both require a capacity of 2 from resource 1. As R1 provides a total capacity of 4 at the beginning of the schedule, J1 and J2 can be processed in parallel.
J3 and J4 in the example do not require any resources. However, both J5 and J6 require a capacity of 2 from R2. Note that during the machine downtime of M2 (interval [6,8)), J5 is paused and does not need resource capacities so that J6 can be processed during this time.
Moreover, machine M2 also requires one unit of resource R2 whenever it is processing a job.

In the following, we provide the full formal specification of the problem.

\subsection{Input Parameters and Variables}


\begin{table}[ht]
    \centering
    \begin{tabular}{p{4cm}l}
    \toprule
    Description & Parameter \\ \midrule
       Set of machines & \( M, \, \vert M \vert = k \) \\
       Set of jobs & \( J, \, \vert J \vert = n \) \\
       Starting dummy jobs & \( b_m \hspace{0.2em} \forall m \in M, J_b = \{b_m \mid m \in M \} \) \\
       Jobs with dummies & \( J_0 = J \cup J_b 
       \) \\
       Eligible machines & \( E_j \subseteq M \quad \forall j \in J_0 \) \\
       Due date of each job & \( d_{j} \in \mathbb{N} \quad \forall j \in J \) \\
       Release date of each job & \( rd_{j} \in \mathbb{N} \quad \forall j \in J \) \\
       Job processing times & \( jp_{j,m} \in \mathbb{N} \quad \forall j \in J_0, m \in M \) \\
       Job setup times & \( s_{m,j_1,j_2} \in \mathbb{N} \hspace{0.2em} \forall m \in M, j_{1/2} \in J_0 \) \\
       Set of resources & \( R, \, \vert R \vert = s \) \\
       Resource capacity & \( rc_{r,t} \in \mathbb{N} \quad \forall r \in R, t \in H \) \\
       Job demands & \( z_{j,r} \in \mathbb{N} \quad \forall j \in J, r \in R \) \\
       Machine demands & \( v_{m,r} \in \mathbb{N} \quad \forall m \in M, r \in R \) \\
       Machine downtimes & \( U = \bigcup_{m \in M} U_m\) \\
       \quad Downtime start & \( us_u \in H \quad \forall u \in U \) \\
       \quad Downtime end & \( ue_u \in H\quad \forall u \in U \) \\
       Set of job precedences & \( P_j=\{(l_p, prec_p)\}  \quad \forall j \in J \) \\
       \quad Precedence time lag & \( l_p \in \mathbb{N} \quad \forall j \in J, p \in P_j \) \\
       \quad Predecessor job & \( prec_p \in J \quad \forall j \in J, p \in P_j \) \\
       Horizon upper bound & \( V \in \mathbb{N} \) \\
       \quad Scheduling horizon & \( H = [0 .. V] \) \\
       \bottomrule
    \end{tabular}
    \caption{Input parameters of the PMSP-PCRCC}
    \label{tab:input_params}
\end{table}

Table~\ref{tab:input_params} summarizes all instance parameters.
We further define the following variables for the PMSP-PCRCC:
\begin{itemize}
    \item Job start times: 
    \eqfontsize
    \( start_j \in \mathbb{N} \quad \forall j \in J_0 \)
    \normalsize
    
    \item Job completion times: 
    \eqfontsize
    \( end_j \in \mathbb{N} \quad \forall j \in J_0 \)
    \normalsize

    \item Indicator variables determining if a job is spanning across a machine downtime:
    \eqfontsize
    \(
     across_{j,u} \in \{0, 1\} \hspace{0.2em} \forall j \in J, 
     u \in U 
    \)
    \normalsize
    
    \item Previous job assignments (capture the direct predecessor of the job on the same machine): 
    \eqfontsize
    \(
     prev_j \in J \cup J_b \setminus \{j\} \hspace{0.2em} \forall j \in J 
    \)
    \normalsize
    
    \item  Machine assignments:
    \eqfontsize
    \( a_j \in E_j \quad \forall j \in J_0 \)
    \normalsize

    \item Total tardiness:
    \eqfontsize
    \( T \in \mathbb{N} \)
    \normalsize
    
    \item Makespan:
    \eqfontsize
    \( C \in \mathbb{N} \)
    \normalsize
    
    \item Total setup time:
    \eqfontsize
    \( S \in \mathbb{N} \)
    \normalsize
\end{itemize}

\subsection{Constraints}
Several constraints impose restrictions on feasible schedules:

\begin{itemize}
    \item All previous job assignments must be different: 
    \eqfontsize
    \begin{equation*}
    prev_{(j_1)} \neq prev_{(j_2)} \quad \forall j_{1/2} \in J 
    \textrm{ where } j_1 \neq j_2
    \end{equation*}
    \normalsize

    \item The starting dummy jobs which represent the machine starts are fixed to the corresponding machines: 
    \eqfontsize
    \begin{equation*}
    a_{(b_m)} = m  \quad \forall m \in M
    \end{equation*}
    \normalsize
    
    \item Set start- and times of starting dummy jobs to 0: 
    \eqfontsize
    \begin{equation*}
    start_j = 0 \land end_j = 0 \quad \forall j \in J_b
    \end{equation*}
    \normalsize

    \item Jobs must be assigned to the same machine as their previous job:
    \eqfontsize
    \(
    a_{(prev_j)} = a_j \quad \forall j \in J 
    \)
    \normalsize
    
    \item Determine if a job is spanning across a machine downtime:
    \eqfontsize
    \begin{gather*}
    (start_j < us_u \land end_j > ue_u) 
    \Leftrightarrow across_{j,u} = 1 \\ \forall u \in U_{(a_j)}, j \in J 
    \end{gather*}
    \normalsize
    
    \item Calculate the job end times, including pauses (note that the job's start time is set to the beginning of the setup time):
    \eqfontsize
    \begin{gather*}
    end_j = start_j + s_{(a_j),(prev_j),j} + jp_{j,(a_j)} \\ + \sum_{u \in U_{(a_j)}} across_{j,u} \cdot (ue_u - us_u) \quad
    \forall j \in J 
    \end{gather*}
    \normalsize

    \item Job start times must be greater than or equal to the release date: 
    \eqfontsize
    \(
    start_j \geq rd_j
    \quad \forall j \in J 
    \)
    \normalsize
    
    \item Job start times must be greater than or equal the previous job's end time:
    \eqfontsize
    \(
    start_j \geq end_{(prev_j)} 
    \quad \forall j \in J 
    \)
    \normalsize
    
    \item Job starts and ends cannot lie within machine downtimes:
    \eqfontsize
    \begin{align*}
    & (start_j < us_u \lor start_j \geq ue_u) \quad \land \\
    & (end_j < us_u \lor end_j \geq ue_u) \quad
    \forall j \in J, 
    u \in U_{(a_j)}
    \end{align*}
    \normalsize   

    \item Sufficient resource capacities must be supplied for all running jobs at any time point (except for paused jobs during machine downtimes):
    \eqfontsize
    \begin{gather*}
    \sum_{j \in J_t}  z_{j,r} + v_{(a_{j}),r} \leq  rc_{r,t} \quad
    \forall t \in H, r \in R \\ J_t = \bigl\{ j \in J | t \in [start_j, end_j] \quad \land \\
    \forall u \in U_{(a_j)}(t \notin [us_u, ue_u]) \bigr\}
    \end{gather*}
    \normalsize

    \item Job precedence constraints must be fulfilled with respect to the precedence time lag:
    \eqfontsize
    \begin{equation*}
     start_j \geq end_{(prec_p)} + l_p \quad \forall j \in J, p \in P_j
    \end{equation*}
    \normalsize

    \item Calculate total tardiness:
    \eqfontsize
    \( T = \sum_{j \in J} \max \{ 0, end_j - d_j \}\)
    \normalsize

    \item Calculate makespan: 
    \eqfontsize
    \( C = \max \{ end_j \mid \forall j \in J \}\)
    \normalsize

    \item Calculate total setup time:
    \eqfontsize
    \( S = \sum_{j \in J} s_{(a_j),(prev_j),j}\)
    \normalsize

\end{itemize}

\subsection{Objective Function}
\label{sec:objective}

The objective function minimizes a weighted sum of the total tardiness, the makespan, and the total setup time (with weights \( w_{1/2/3} \in \mathbb{R} \)):
\eqfontsize
\(
    \quad minimize \quad w_1 \cdot T + w_2 \cdot C + w_3 \cdot S
\)
\normalsize


\section{Constraint Modeling Approach}
\label{sec:cp_model}

This section provides a constraint model using the solver-independent modeling language MiniZinc~\cite{nethercote_minizinc_2007}. Thus, the model can be used with state-of-the-art MIP- and CP-solving technology as an exact solution method for the PMSP-PCRCC.
Some parts of the model directly represent the problem definition in Section~\ref{sec:problem-def}. Therefore, we only describe the key differences and the global constraints in the following.\footnote{The full MiniZinc model is provided as supplementary material and will be made publicly available.} 

Unlike the specification in Section~\ref{sec:problem-def}, we specify the resource capacities provided by a time interval list $RList$, which consists of resource tuples $(r, s, e, c)$ where $r \in R$ (=resource), $s < e \in H$ (=start/end time) and $c \in \mathbb{N}$ (=capacity).
The set of intervals $[s, e]$ is assumed to cover the entire scheduling horizon $H$ for each resource.
We further extend the input parameters by creating a set $J_R$ of additional placeholder jobs with fixed start- and end times to efficiently capture the availability of resources as follows:
For each resource $r \in R$, we first determine the provided maximum capacity ($cMax_r$) throughout the scheduling horizon $H$.
Then we create placeholder jobs $j_i \in J_R$ for each tuple $\{(r_i, s_i, e_i, c_i) \in RList | c_i < cMax_{r_i}\}$  where the provided capacity is less than the maximum (the placeholder job will have a resource demand equal to the difference to the maximum capacity) and set the start and end times of the job accordingly (\(start_{j_i} = s_i \land end_{j_i} = e_i\)).

Similarly, as in Section~\ref{sec:problem-def}, we use decision variables for each job $j$ that capture the start ($start_j$) and the end time ($end_j$), as well as the machine assignment ($a_j$).
Furthermore, we use auxiliary variables $jobStartPeriod_j$ and $jobCompletionPeriod_j$ for each job $j$ to encode the index of the selected available machine periods, that is, of the time intervals between machine downtimes, in which each job starts and ends.
Since jobs can be paused during machine downtimes and no resources are consumed during pauses, we further model the individual uninterrupted parts of each job.
Thus, we introduce additional auxiliary variables that encode the durations of each potential job part within every machine period: $startjobParts(j,p)$ and $durationJobParts(j,p)$ for all jobs $j$ in $J^* = J \cup J_R$ and $p \in \mathcal{P}$ where $\mathcal{P}$ is the index set of machine periods.

These variables are also used for the placeholder jobs in $J_R$, allowing us to precisely capture the capacity of resources $r \in R$ consumed by each job $j \in J^*$ during all machine periods $p \in \mathcal{P}$ by $SecResourceUsage(r,j,p)$ (which is defined by the sum of the demand for jobs $z_{j,r}$ and the demand for machines $v_{m,r}$, considering $a_j=m$ and job $j$ is running in period $p$).
Thus, start time, duration, and resource usage are fixed for each placeholder job $j_i \in J_R$ using the parameters from each tuple in $RList$ to ensure that the available resource capacity for normal jobs is $c_i$ (setting the machine period index for the placeholder jobs to 1): 
\eqfontsize
\begin{align*}
    startJobParts(j_i,1) & =  s_i \\ 
    durationJobParts(j_i,1) & = e_i-s_i\\
    secResourceUsage(r_i, j_i,1)& =cMax_{r_i}-c
\end{align*}
\normalsize

Finally, we model the resource requirements using the global constraint \texttt{cumulative}~\cite{schutt_explaining_2011}, as state-of-the-art constraint-solving technology provides efficient algorithms to handle this constraint.
Given the set $J^*$ that includes all normal jobs and placeholder jobs, we model the resource requirements for each resource $r \in R$ as follows:
\eqfontsize
\begin{align*}
&\texttt{cumulative}(S, D, R, c), \text{ where } \\&S = (startJobParts(j,p))_{j \in J^*, p \in \mathcal{P}}, \\
&D = (durationJobParts(j,p))_{j \in J^*, p \in \mathcal{P}}, \\
&R= (secResourceUsage(r,j,p))_{j \in J^*, p \in \mathcal{P}}, \\
&c = cMax_r. 
\end{align*}
\normalsize
The global constraint $\texttt{cumulative}(S, D, R, c)$ ensures that a set of jobs with start times $S$, durations $D$ and resource demands $R$ never require more than the global resource capacity $c$ at any time.

\section{Construction Heuristic}
\label{sec:construction_heuristic}

In the following, we introduce a construction heuristic that aims to produce efficient schedules for large-scale instances of the PMSP-PCRCC, which cannot be tackled with exact techniques.
Furthermore, the method can quickly provide efficient initial solutions for the metaheuristic based on local search that we propose in Section~\ref{sec:sim-ann}.

The core of the method uses a dispatching rule that starts from an empty solution and schedules all jobs iteratively based on a set of rules that aims to optimize the makespan and tardiness objectives.
In each iteration, the heuristic algorithm determines a set of \emph{active} jobs $A\subseteq J$, which can be scheduled next without violating any job precedence constraints (i.e., jobs for which all required predecessor jobs are already scheduled). 
The set $A$ is then sorted and partitioned according to the jobs' due dates. 
Let $A_{d^{*}}\subseteq A$ be the subset of active jobs with the \emph{earliest} due date~$d^{*}$. 

The algorithm then calculates for each job $j^{*}\in A_{d^{*}}$ and each feasible machine assignment $m^{*}$ the earliest possible completion time of the job. All relevant constraints regarding release dates, resource availability, etc. are considered when calculating these completion times.
The job with the lowest possible completion time is then scheduled as early as possible (ties are broken by the lowest job index according to the input parameters).
At the end of each iteration, $A$ is updated by removing $j^{*}$ and adding new jobs that become active.

The algorithm ends once all jobs are scheduled. 
Algorithm~\ref{alg:constructive} presents the main routine of the proposed construction heuristic.

\begin{algorithm}[ht]
\small
\caption{Construction Heuristic for the PMSP-PCRCC}
\label{alg:constructive}
\textbf{Input:} Problem Instance\\
\textbf{Output:} Solution schedule
\begin{algorithmic}[1] 
\STATE $A\leftarrow\{j\in J\mid P_j=\emptyset \}$
\WHILE{$A\not=\emptyset$}
\STATE $A_{d^{*}}\leftarrow\{j\in A\mid d_j=d^{*}\}$ with $d^{*}=\min_{j\in A} d_j$
\STATE select job $j^{*}$ and machine $m^{*}\in E_{j^{*}}$ from $A_{d^{*}}$ that has the earliest possible completion time
\STATE schedule $j^{*}$ on machine $m^{*}$ as early as possible
\STATE remove $j^{*}$ from $A$ and add new active jobs
\ENDWHILE
\RETURN solution schedule
\end{algorithmic}
\end{algorithm}

The construction heuristic guarantees that all job precedences are fulfilled in the resulting schedule.
However, the routine might reach a state where not all jobs can be scheduled as no more resources are available due to the calendar restrictions.
In such a case, the algorithm simply ignores the resource requirements and continues to schedule the remaining jobs at the end of the scheduling horizon.
The resulting solution will be infeasible, but can still serve as an initial solution for the metaheuristic approach we propose in the next section, that can then potentially repair the violations. 

\section{Metaheuristic Approach}
\label{sec:sim-ann}

In this section, we propose a metaheuristic based on local search and simulated annealing that can further improve the initial schedules generated by the construction heuristic from Section~\ref{sec:construction_heuristic} by iteratively applying neighborhood moves on a current solution.
The approach uses an innovative permutation-based solution representation, which we describe next in Section~\ref{sec:perm_sol_rep}.
Afterward, in Section~\ref{sec:neighborhood}, we describe the generation of neighborhood moves before we present the move acceptance criteria in Section~\ref{sec:sa_acceptance}.

\subsection{Solution Representation}\label{sec:perm_sol_rep}
Previous work on local search methods for related variants of the PMSP (e.g., in~\cite{moser_exact_2022}) has proposed to represent solutions efficiently using a list of jobs for each machine instead of storing the start and end times for each job.
Such a representation suffices as long as it is reasonable to schedule each job based on the list orderings as early as possible after its predecessor has ended.
However, in our case, the job ordering in such a representation becomes unclear, as predecessor constraints and resource constraints can have interleaving effects between the jobs on different machines. In other words, in many situations, it could make sense that a job waits for the completion of a predecessor job (or a job that uses a shared resource) that is scheduled on another machine.
Thus, we propose to use a permutation-based solution representation for the PMSP-PCRCC which we describe in the following.

\begin{figure}[ht]
\centering
\begin{subfigure}[b]{0.9\columnwidth}
   \includegraphics[width=0.95\columnwidth]{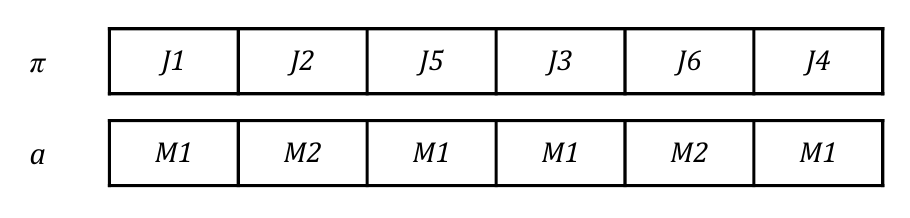}
   \caption{Visualized solution representation $(\pi, a)$ for the example depicted in Figure~\ref{fig:example_schedule}.}
   \label{fig:example_sol_rep}
\end{subfigure}
\begin{subfigure}[b]{0.9\columnwidth}
   \includegraphics[width=0.95\columnwidth]{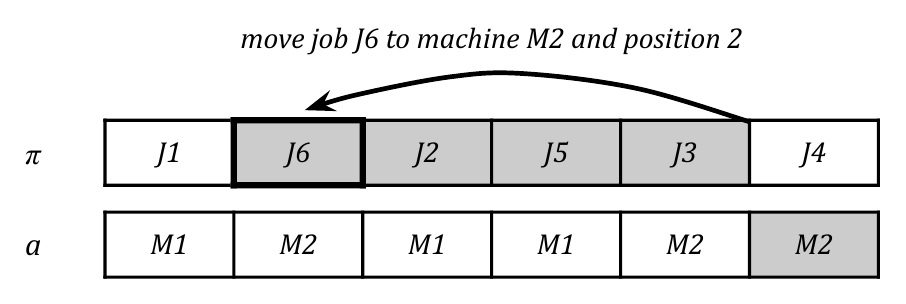}
   \caption{Solution obtained after applying the move that shifts job $J6$ to machine $M2$ and to position $2$ in the global schedule.}
   \label{fig:example_move}
\end{subfigure}
\caption{Visualization of a solution representation and neighborhood move  for the example shown in Figure~\ref{fig:example_schedule}.}
\end{figure}

Instead of storing job orders in separate lists for each machine, we represent solutions by using a global job ordering in a single list and storing machine assignments for each job in a second list.
Intuitively, the jobs' start times of any given solution imply a total ordering of the jobs. 
Thus, a global order, that is, a permutation $\pi=(\pi_i)_{i=1,\dots,n}$ of the jobs in $J$ together with the list of machine assignments $a = (a_j)_{j\in J}$ can be decoded into a candidate solution by iteratively scheduling the jobs in the given order on the specified machines as early as possible.
As an example, the solution representation $(\pi, a)$ for the example given in Figure~\ref{fig:example_schedule} is visualized in Figure~\ref{fig:example_sol_rep}.

To determine the concrete start times for each job from a given solution representation, we use a dispatching rule that keeps track of resource usage and calendar information and determines the earliest start time where all resource requirements of the job are fulfilled.
The dispatching routine can get stuck if no more resources or machines are available at the end of the schedule horizon, leading to an infeasible solution state.
However, we can argue that if a feasible solution exists for an instance, there must also exist a solution representation that is decoded into the optimal solution, as all possible permutations can be stated, and it is never reasonable to delay a job if the job order is fixed.
Similarly, the job precedence constraints can be violated by the global job order leading to an infeasible schedule.
However, the precedence requirements can be checked simply by verifying the total order of all jobs involved in precedences. 

\subsection{Neighborhood Exploration}
\label{sec:neighborhood}

In the proposed metaheuristic, we use a neighborhood operator that selects a single job and changes its machine assignment and/or its position in the global job order.
A possible neighborhood move for the example schedule in Figure~\ref{fig:example_schedule} is visualized in Figure~\ref{fig:example_move}. Job $J6$ is moved from position 5 in the global job order to position 2 and the machine assignment changes from $M1$ to $M2$.
Note that moving $J6$ to position 2 also increments the positions of jobs $J2$, $J3$ and $J5$.  

Given a candidate solution, a large number of possible neighborhood moves can be generated.
As evaluating potential cost improvements for all possible neighbors can quickly become computationally expensive, especially for large practical instances, we propose to randomly sample only one neighborhood move in each local search iteration.
Furthermore, we only consider neighborhood moves that will not violate any job precedence or machine eligibility constraints.
Thus, our neighborhood generation procedure uniformly samples a single neighbor from the set of all possible neighbors that do not violate job precedences or machine eligibility.
Since the initial solution generated by the construction heuristic guarantees that there are no violations regarding these constraints, the metaheuristic never reaches a state where these requirements are unfulfilled.



\subsection{Cost Function and Move Acceptance}
\label{sec:sa_acceptance}
The costs of a candidate solution are defined by the sum of the multi-objective value defined in Section~\ref{sec:objective} and an additional hard constraint violation count that is multiplied by a big M value. The big M value is chosen based on the upper bounds to the tardiness-, setup time-, and makespan objectives so that a single hard constraint violation causes incomparably worse costs than any feasible solution's cost.
As mentioned in sections~\ref{sec:construction_heuristic} and~\ref{sec:neighborhood}, due to the design of the initial solution generation with the construction heuristic and the neighborhood operator, the only constraint violations that may occur during local search are resource violations or machine availability violations. The violation count is calculated by summation of the number of jobs with unfulfilled resource constraints and the number of jobs with unfulfilled machine availability constraints.

At the end of each local search iteration, the potential change in this cost function caused by the randomly selected neighborhood move is evaluated.
In case of a cost improvement or if the cost stays the same, the metaheuristic always applies the candidate move and proceeds with the next iteration.
If the move leads to an increase in cost, a simulated annealing-based acceptance function determines whether the move is applied.

Simulated annealing~\cite{kirkpatrick_optimization_1983} (SA) is a metaheuristic technique that has been successfully used on many related problems to escape local optima during search.
The SA acceptance function determines whether a worsening move is accepted based on the change in costs and a temperature value.
At the beginning of the search algorithm, the temperature value is set to an initial value that is set by a parameter.
After each search iteration, the temperature is reduced based on a geometrical cooling scheme, where the current temperature is decreased by multiplication with a cooling rate value. 

We use a variant of SA that dynamically adjusts the cooling rate according to the remaining run time so that a minimum temperature value (which is provided as a parameter) is reached exactly at the end of a given run-time limit.
The following formula is used to calculate the cooling rate $c_i$, where $x$ denotes the number of cooling steps left (which is determined by dividing the remaining time by the average runtime for all previous search iterations), $T_{min}$ specifies the minimum temperature and $t_c$ the current temperature: \(c_i := \sqrt[x]{T_{min}/t_c} \).
Finally, we calculate the acceptance probability $p$ of a worsening move depending on the change in the cost function $\delta$ and the current temperature $t_c$ as \( p := \exp(-\delta/t_c)\).


\section{Experimental Evaluation}
\label{sec:experiments}

In this section, we first describe the set of benchmark instances and then present the random instance generation method we have developed to generate random instances.
Afterward, we provide details about the computational environment and parameter configuration of the metaheuristic before we present and discuss all computational results.


\subsection{Benchmark Instances}
\label{sec:instances}


We used two sets of instances for our empirical evaluation: A set of real-life instances from the packaging industry and a large set of randomly generated instances.
Regarding the objective function, we chose uniform objective weight settings (\(w_{1/2/3} = 1\)) as the impact of the individual objective cost components lies within similar magnitudes for the real-life application from which our problem originates.

\subsubsection{Real-life Instances}
\label{sec:random_generator}
The set of realistic instances was extracted directly from real-life industrial data.  
The set comprises 25 diverse instances, featuring up to 685 jobs, between 3 and 63 machines, and 3 to 80 resources.
Detailed instance size parameters are presented in the first three columns of each row in Table~\ref{tab:main_results_real_world}, where $n$ is the number of jobs, $k$ is the number of machines, and $s$ is the number of resources.

\subsubsection{Random Instances}
We developed a random instance generator inspired by instance generation algorithms for PMSPs proposed in previous work~\cite{vallada_genetic_2011, moser_exact_2022}.
The extended generator utilizes two main concepts to build random instances: Job materials and a reference solution.

A single material is randomly assigned to each job to build classes of jobs using the same material. Jobs within the same material class then share the same required resources but may require varying amounts of capacity. Furthermore, no setup time is necessary between jobs within the same class, while setup times between different materials are chosen randomly.

After the processing times, setup times, and resource consumption of the jobs have been determined, a reference solution is created by a dispatching rule that processes the jobs in random order and aims to assign jobs to a machine that leads to the shortest setup times.
The reference solution is then used to select reasonable values for the due- and release dates, to sample chains of job precedences, and to generate machine downtimes.
To create meaningful precedence relations, we create each precedence chain by randomly selecting two to four jobs that are non-overlapping in the reference solution.
The order of these jobs in the reference solution then implies the precedence relations. Based on the resource requirements of the jobs and machines in the reference solution, a resource calendar is created. The calendar is partitioned into resource periods by breaks in resource requirements. The largest cumulative requirement within such a period then establishes the resource capacity for this resource and period.

Using the random generator, we created a set of random instances that consists of 10 or 20 jobs (\(n \in \{10, 20\}\)).
The set of generated instances is further organized into eight categories using different instance size configurations, where each category consists of 20 instances.
The instance size parameters used to configure the random instance generator consist of the number of precedence chains (\(c \in \{\frac{n}{10}, \frac{n}{4}\}\)), the number of machines (\(k \in \{2,5\}\)), and the number of resources (\( s \in \{5, 10\}\)).


The full details and source code of the random instance generator, as well as all used benchmark instances, will be made publicly available.




\subsection{Computational Environment}
All experiments, as well as parameter tuning, were executed on a computing cluster with 13 nodes, each featuring two Intel Xeon E5-2650 v4 CPUs (12 cores @ 2.20GHz).

We further implemented the constraint modeling approach using MiniZinc v2.8.5~\cite{nethercote_minizinc_2007} together with solver backends for the Gurobi MIP solver v10.0.1~\cite{gurobi} and the CP-SAT CP solver v9.10~\cite{cpsatlp}. To support Gurobi, MiniZinc automatically transforms the model into a MIP model~\cite{belov_improved_2016}.
For both solvers, we set a runtime limit of 1 hour and conducted a single run per instance.

The construction heuristic and the metaheuristic were both implemented using the C\# programming language.
We performed a single experimental run for each instance using the construction heuristic, which was completed within short runtimes for all instances.
As the metaheuristic uses a probability-based acceptance function, we performed 12 repeated runs for each problem instance, where each individual had a runtime limit of 5 minutes. We report the best results out of 12 runs in the final results, to allow a reasonable comparison with the exact results reached within one hour.


\subsection{Algorithm Configuration}

For the move acceptance function of the proposed metaheuristic, efficient temperature parameters must be configured.
Thus, we used the automated parameter configuration tool irace~\cite{irace} for parameter tuning.

We conducted two separate tuning runs for the random instances and the real-life instances.
As a training set for the first tuning run, we generated an additional 80 random instances using the same instance size parameters as in Section~\ref{sec:random_generator}, but this time, we generated 10 instances for each category. 
The second training set, which was used for the real-life instances tuning run, also consists of 80 additionally generated instances with similar parameters, except that the number of jobs was set to 100, to focus on larger instances.

Both tuning runs carried out a total of 10000 experiments, starting from an initial parameter configuration that uses an initial temperature of 1000 and a minimum temperature of 0.0001 which was determined using preliminary manual tuning trials. Each tuning experiment was performed with a runtime of 5 minutes.
Table~\ref{tab:parameter_tuning_options} summarizes the possible parameter options that were explored by the automated algorithm configuration tool, as well as the resulting parameter values.
\begin{table}[ht]
\centering
\fontsize{9pt}{10pt}\selectfont
\begin{tabular}{p{1.5cm}p{3.2cm}l}
\toprule
                & Initial Temp.                   & Minimum Temp. $T_{min}$           \\ \midrule
Options & 10, 50, 100, \dots , 1500 & 0.01, $10^{-3}$, $10^{-4}$, $10^{-5}$ \\ \cmidrule{1-3}
 Random            & 600                                   & 0.001                        \\
 Real-life          & 500                                   & 0.001                        \\ \bottomrule
\end{tabular}
\caption{Tuning results for the temperate parameters.}
\label{tab:parameter_tuning_options}
\end{table}

\subsection{Computational Results}

\begin{figure}[tb]
    \centering
    \includegraphics[width=0.9\columnwidth]{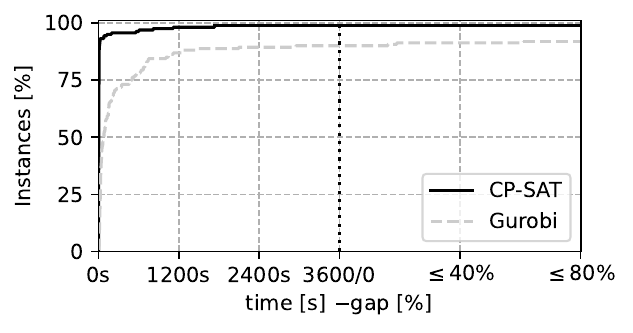}
    \caption{Performance plots for instances with up to twenty jobs solved with CP-SAT and Gurobi. Time limit: one hour.}
    \label{fig:performance_plot}
\end{figure}

\begin{table*}[tb]
\fontsize{9pt}{10pt}\selectfont
\centering
\begin{tabular}{rrrrrrrrrrrrr}
\toprule
   &    &    &\multicolumn{4}{c}{CP-SAT} & \multicolumn{5}{c}{SA} \\ \cmidrule(lr){4-7}\cmidrule(lr){8-12}
c & k & s &   $\overline{C}$ &  $\overline{T}$ &  $\overline{S}$ &        \#opt &  $\overline{C}$ &  $\overline{T}$ &  $\overline{S}$ &        \#opt & \#no sol\\
\midrule
\multirow{4}{*}{\(\frac{n}{10}\)} & \multirow{2}{*}{2} & 5  &   \textbf{769.4} &  \textbf{460.9} &  \textbf{272.8} &  \textbf{16} &  \textbf{769.4} &  \textbf{460.9} &  \textbf{272.8} &  \textbf{16} &             0 \\
  &   & 10 &  \textbf{1275.3} &  \textbf{382.6} &           655.8 &  \textbf{16} &          1275.8 &           435.5 &  \textbf{631.3} &           10 &             4 \\
\cmidrule{2-12}
  & \multirow{2}{*}{5} & 5  &   \textbf{509.5} &  \textbf{577.1} &  \textbf{234.4} &  \textbf{15} &  \textbf{509.5} &  \textbf{577.1} &  \textbf{234.4} &           14 &             1 \\
  &   & 10 &   \textbf{506.2} &  \textbf{538.9} &  \textbf{261.6} &  \textbf{13} &  \textbf{506.2} &           546.5 &           270.2 &           12 &             0 \\
\cmidrule{1-12}
\cmidrule{2-12}
\multirow{4}{*}{\(\frac{n}{4}\)} & \multirow{2}{*}{2} & 5  &   \textbf{769.8} &  \textbf{811.5} &  \textbf{366.1} &  \textbf{13} &  \textbf{769.8} &  \textbf{811.5} &  \textbf{366.1} &           12 &             1 \\
  &   & 10 &   \textbf{801.2} &  \textbf{206.9} &  \textbf{441.7} &  \textbf{16} &  \textbf{801.2} &  \textbf{206.9} &  \textbf{441.7} &           15 &             1 \\
\cmidrule{2-12}
  & \multirow{2}{*}{5} & 5  &   \textbf{545.9} &  \textbf{367.6} &  \textbf{313.3} &  \textbf{17} &  \textbf{545.9} &  \textbf{367.6} &  \textbf{313.3} &           16 &             1 \\
  &   & 10 &   \textbf{465.3} &  \textbf{152.4} &  \textbf{244.3} &  \textbf{12} &  \textbf{465.3} &  \textbf{152.4} &  \textbf{244.3} &  \textbf{12} &             0 \\
\bottomrule
\end{tabular}
\caption{Aggregated results obtained by CP-SAT and SA for instances with up to twenty jobs that could be solved to proven optimality by CP-SAT\@.}
\label{tab:cpsat_sa_results}
\end{table*}

\begin{table*}[!tb]
\fontsize{9pt}{10pt}\selectfont
\centering
\begin{tabular}{rrrrrrrrrrr}
\toprule
 & & & \multicolumn{4}{c}{CA} & \multicolumn{4}{c}{SA} \\ \cmidrule(lr){4-7}\cmidrule(lr){8-11}
n & k & s  &            $C$ &             $T$ &            $S$ & $|J_{\dag}|$ &             $C$ &              $T$ &            $S$ &  $|J_{\dag}|$ \\
\midrule
2   & 3  & 3  &   \textbf{165} &      \textbf{0} &     \textbf{0} &   \textbf{0} &    \textbf{165} &       \textbf{0} &     \textbf{0} &    \textbf{0} \\
15  & 10 & 5  &  \textbf{8230} &  \textbf{15165} &    \textbf{30} &   \textbf{2} &   \textbf{8230} &   \textbf{15165} &    \textbf{30} &    \textbf{2} \\
16  & 14 & 9  &  \textbf{8767} &  \textbf{12791} &             34 &   \textbf{6} &   \textbf{8767} &            12792 &    \textbf{32} &    \textbf{6} \\
19  & 14 & 6  &   \textbf{230} &    \textbf{169} &             38 &   \textbf{2} &    \textbf{230} &     \textbf{169} &    \textbf{37} &    \textbf{2} \\
21  & 19 & 7  &   \textbf{913} &             392 &             38 &  \textbf{11} &    \textbf{913} &     \textbf{389} &    \textbf{37} &   \textbf{11} \\
31  & 19 & 9  &  \textbf{1580} &            2987 &             63 &            5 &   \textbf{1580} &    \textbf{1449} &    \textbf{45} &    \textbf{3} \\
107 & 39 & 25 &            250 &    \textbf{309} &            504 &   \textbf{7} &    \textbf{223} &     \textbf{309} &   \textbf{386} &    \textbf{7} \\
159 & 42 & 19 &   \textbf{805} &    \textbf{424} &            719 &   \textbf{6} &    \textbf{805} &     \textbf{424} &   \textbf{512} &    \textbf{6} \\
166 & 37 & 26 &            807 &    \textbf{485} &            803 &  \textbf{15} &    \textbf{421} &              563 &   \textbf{562} &   \textbf{15} \\
255 & 36 & 28 &           3406 &           26151 &           1083 &           56 &   \textbf{3397} &   \textbf{11941} &   \textbf{865} &   \textbf{35} \\
279 & 47 & 28 &           1019 &           11188 &           1027 &           87 &   \textbf{1007} &    \textbf{6518} &   \textbf{811} &   \textbf{67} \\
298 & 52 & 47 &           2254 &           20542 &           1421 &           36 &   \textbf{2058} &    \textbf{3798} &  \textbf{1086} &   \textbf{21} \\
321 & 59 & 50 &           6397 &          187477 &           1077 &           65 &   \textbf{6081} &   \textbf{30715} &  \textbf{1000} &   \textbf{24} \\
340 & 51 & 53 &  \textbf{5147} &           28306 &           1896 &           59 &   \textbf{5147} &   \textbf{14851} &  \textbf{1503} &   \textbf{36} \\
352 & 51 & 53 &           1589 &           11309 &           1701 &           43 &   \textbf{1389} &    \textbf{5086} &  \textbf{1447} &   \textbf{30} \\
375 & 40 & 36 &           6482 &           21871 &           1756 &           82 &   \textbf{6454} &   \textbf{11993} &  \textbf{1494} &   \textbf{53} \\
420 & 39 & 40 &           6963 &          158770 &           1850 &           90 &   \textbf{6083} &   \textbf{29701} &  \textbf{1499} &   \textbf{59} \\
426 & 48 & 59 &           6188 &          222006 &           2039 &          113 &   \textbf{6180} &   \textbf{46574} &  \textbf{1871} &   \textbf{45} \\
432 & 52 & 46 &           5551 &          166609 &           1893 &          114 &   \textbf{5136} &   \textbf{36333} &  \textbf{1684} &   \textbf{75} \\
442 & 52 & 49 &           5024 &          110181 &           1709 &           45 &   \textbf{4728} &    \textbf{9705} &  \textbf{1547} &   \textbf{18} \\
476 & 63 & 69 &           8382 &          334166 &           2078 &           89 &   \textbf{7621} &   \textbf{86586} &  \textbf{2052} &   \textbf{66} \\
564 & 59 & 79 &          12195 &          484415 &  \textbf{2690} &          175 &  \textbf{12132} &  \textbf{253142} &           2758 &  \textbf{144} \\
620 & 61 & 80 &          12157 &          637144 &  \textbf{3160} &          325 &  \textbf{11646} &  \textbf{264348} &           3271 &  \textbf{135} \\
653 & 60 & 70 &          10123 &          545212 &  \textbf{3002} &          376 &   \textbf{9074} &  \textbf{212169} &           3055 &  \textbf{171} \\
685 & 61 & 74 &          11012 &         1475112 &  \textbf{3045} &          330 &  \textbf{10424} &  \textbf{728056} &           3125 &  \textbf{247} \\
\bottomrule
\end{tabular}
\caption{Main results of Constructive Algorithm (CA), and the metaheuristic (SA) for real-life instances.}
\label{tab:main_results_real_world}
\end{table*}

First, we investigate the ability of the proposed exact approach to solve instances to proven optimality.
Figure~\ref{fig:performance_plot} shows the result obtained with CP-SAT and Gurobi as a performance plot for randomly generated instances with ten jobs. 
The left part of the plot shows the number of instances that could be solved to proven optimality or for which unsatisfiability could be proven. 
The right part shows the remaining optimality gap for instances where a solution could be found within the one-hour time limit. 
The results show that CP-SAT can solve more instances to proven optimality or unsatisfiability in less running time than Gurobi in our experiments.

Note that the random instances are not necessarily satisfiable, meaning that it might not be possible to schedule all jobs without violating machine- or resource availabilities. 
This scenario can occur in practice, for example, if the planning horizon is too short or it is of interest whether a certain number of jobs can still be planned within a given time frame.
In such cases, the exact approach can prove that the instance is unsatisfiable. 
In particular, CP-SAT can prove optimality in 118 cases and unsatisfiability in 40 cases out of 160 instances. 

Next, we want to investigate the ability of the proposed metaheuristic approach (SA) to reach optimal solutions. 
To this end, Table~\ref{tab:cpsat_sa_results} considers only instances that are solved to proven optimality by CP-SAT\@.
The first three columns display the size parameters of the different randomly generated instance categories (as explained in Section~\ref{sec:instances}), whereas the remaining columns show aggregated results per instance category.
The columns \#opt report the number of instances that could be solved to proven optimality in the case of CP-SAT. In the case of SA, the column \#opt lists the number of times SA finds a solution with the same objective values as CP-SAT\@.
Furthermore, SA reports in column \#no solution found also the number of times when CP-SAT finds an optimal solution, but SA cannot find a feasible solution.
The columns $\overline{C}$, $\overline{T}$, and $\overline{S}$ show the averaged values of the makespan, total tardiness, and total setup time, respectively, over instances that are solved to proven optimality by CP-SAT and for which SA found a feasible solution. 
Overall, SA finds optimal solutions in 102 out of 118 cases, and only in eight cases, SA finds no feasible solution.
Thus, the results show that SA could successfully achieve optimal results in the majority of cases.


\begin{figure}[t]
    \centering
    \includegraphics[width=0.9\columnwidth]{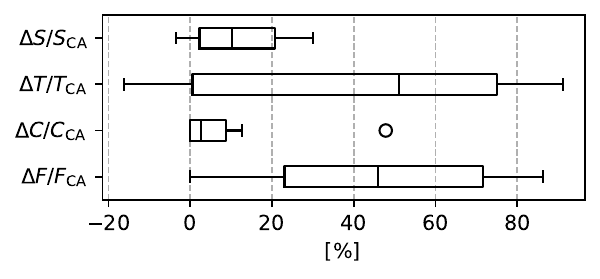}
    \caption{Relative improvements of objective components obtained by SA with respect to the results obtained by CA, averaged over all real-life instances.}
    \label{fig:box_plots}
\end{figure}


Finally, Table~\ref{tab:main_results_real_world} shows the main results obtained by the proposed construction heuristic (CA) and SA for real-life instances. 
Here, the exact methods could only find a feasible solution for the smallest instance within the runtime limit, where it reached the same results as the heuristic.
However, SA could successfully solve all listed instances.
Columns $C$, $T$, and $S$ report the obtained values for makespan, total tardiness, and total setup time, respectively. 
Columns $|J_\dag|$ report the number of tardy jobs, i.e., jobs that do not meet their due date. 
The relative improvement made by SA w.r.t.\ the results obtained by CA are shown in Figure~\ref{fig:box_plots}, for the individual objective components and the aggregated objective $F$.
The relative improvement $\Delta X/X_{\mathrm{CA}}$ with $X\in\{C, T, S, F\}$ is computed by $1 - X_{\mathrm{SA}}/X_{\mathrm{CA}}$, where $X$ is a placeholder for the corresponding objective components.
Overall, the average relative improvement w.r.t.\ the aggregated objective is $41.6\%$. 
Broken down into the individual objective components, the average relative improvement is $5.7\%$, $44.5\%$, and $11.8\%$ for the makespan, total tardiness, and total setup time, respectively. 
On average, SA can reduce the number of tardy jobs by $34.44$ jobs.
The results show that SA could significantly improve the solution quality compared to CA in most cases.

\section{Conclusion}

We introduced and formally defined the PMSP-PCRCC, a novel variant of the PMSP that involves complex job precedences and cumulative resource constraints with calendars. In addition, we provided a collection of benchmark instances that include real-life scheduling scenarios from the packaging industry as well as a large set of random instances.

We further proposed an exact approach using constraint modeling and a metaheuristic approach to solve the problem.
The experimental results showed that the exact method could successfully prove optimality for most small instances.
However, for large real-life instances, only the proposed heuristic techniques were able to provide high-quality solutions in a reasonable time.
In particular, the metaheuristic technique using local search and simulated annealing could overall provide the best results for realistic scenarios in our experiments. In addition, the metaheuristic reached optimal results for the large majority of small-sized instances.
Future work could investigate a hybridization of the exact and heuristic techniques within the framework of a large neighborhood search.




\bibliography{references}

\appendix

\end{document}